\documentclass[journal]{IEEEtran}
\usepackage{cite}               
\usepackage{amsmath,amssymb,amsfonts} 
\usepackage{graphicx}           
\usepackage{textcomp}           
\usepackage{xcolor}             
\usepackage{booktabs}           
\usepackage{float}              
\usepackage{url}                
\usepackage[hidelinks]{hyperref} 

\begin{document}
%
\title{Product-Aware Deep Autoencoders for Robust Process Monitoring in Multi-Product Cyber-Physical Systems}
%
%
%

\author{MD Shafikul Islam, Jordan Carden}

\maketitle

\begin{abstract}
As Industry 4.0 accelerates the integration of Cyber-Physical Systems (CPS) in manufacturing, robust anomaly detection has become critical for ensuring process safety and security. Current data-driven approaches typically employ "product-agnostic" or global models trained on the aggregate of all normal operating data. However, modern industrial facilities frequently operate under diverse product grades or operational modes, creating non-stationary data distributions. While computationally simple, these global models inherently expand their decision boundaries to accommodate the variance of multiple modes, creating a "blind spot" where subtle anomalies or targeted cyber-physical attacks may be masked by the wide acceptance region of the model. In this work, we first demonstrate that the vulnerability described above is present in global-agnostic models operating across multiple product grades. We then present a Product-Aware Autoencoder as a principled mitigation that restricts the learning domain to grade-specific distributions. While this approach reduces the identified blind-spot risk, we do not claim it as the optimal mitigation among all possible alternatives. We rigorously validate this approach against a Global Agnostic baseline using the Extended Tennessee Eastman Process (TEP) benchmark. Our empirical results indicate that the Product-Aware framework performs comparably to the global baseline on standard detection metrics, while offering improved robustness to product-grade-specific operating modes. Most critically, stress tests simulating our hypothetical attack scenarios reveal that while the global model fails to detect operational deviations in 77.8\% of the scenarios, the product-aware system achieves 100\% detection accuracy. These findings suggest that, in flexible manufacturing environments, generalized anomaly detectors can pose non-trivial security risks, motivating a shift toward mode-aware diagnostic architectures.

\end{abstract}

\begin{IEEEkeywords}
Anomaly Detection, Cyber-Physical Systems, Deep Autoencoders, Multi-Mode Process Monitoring, Tennessee Eastman Process, Industrial Control Systems Security.
\end{IEEEkeywords}

%
\IEEEpeerreviewmaketitle

\section{Introduction}

The landscape of modern process industries, particularly within the chemical and petrochemical sectors, has undergone a profound transformation. Driven by the paradigms of Industry 4.0, these sectors have embraced increased automation, interconnectivity, and the deployment of Cyber-Physical Systems (CPS). While these advancements have optimized production throughput and flexibility, they have inadvertently expanded the attack surface of critical infrastructure \cite{iaiani2021analysis}. The integration of Information Technology (IT) with Operational Technology (OT) implies that industrial control systems (ICS) are no longer isolated air-gapped islands, but are increasingly exposed to sophisticated cyber threats. In this high-stakes environment, where physical safety, environmental protection, and economic continuity intersect, the capability to detect anomalies rapidly and accurately is paramount.

The vulnerability of these systems is not merely theoretical. High-profile incidents have underscored the severity of the threat landscape. For instance, in 2017, the Triton malware infiltrated the Safety Instrumented System (SIS) of a petrochemical plant, specifically targeting Schneider Electric’s Triconex controllers with the intent to cause equipment malfunctions or disable safety protocols \cite{alladi2020industrial}. Similarly, the 2014 cyberattack on a German steel mill demonstrated that digital intrusions could transcend data theft and cause catastrophic physical damage to massive industrial equipment. These realities highlight a critical need for vigilant monitoring systems capable of identifying malicious interventions or abnormal process behaviors before they cascade into hazardous situations \cite{abshari2025survey}.

However, securing these facilities is complicated by the inherent complexity of modern manufacturing. Unlike static systems that operate under a single steady state, modern plants are dynamic, multi-product environments. To maximize efficiency, facilities frequently switch between different operational modes, altering feedstocks, changing product grades such as polymer density, or adjusting load levels. Consequently, the statistical characteristics of the process data are non-stationary; what is considered "normal" behavior in one production mode may be flagged as an anomaly in another \cite{quinones2019data}. Traditional multivariate monitoring techniques often assume a single nominal operating regime. When applied to a process that transitions between multiple operation regions, these "product-agnostic" models suffer from severe performance degradation \cite{wang2012process}. They may generate excessive false alarms during legitimate mode transitions or, conversely, fail to detect subtle attacks that are masked by the variance of a different operating mode.

To address the limitations of traditional statistical methods, the research community has increasingly turned toward data-driven approaches, particularly deep learning \cite{alfeo2020using}. Among these, Autoencoder (AE) neural networks have emerged as a powerful tool for process monitoring. By learning to compress and reconstruct high-dimensional sensor data, autoencoders can capture complex, non-linear correlations that linear methods like Principal Component Analysis (PCA) often miss \cite{xiao2023fault}. In a standard setup, a significant increase in the network's reconstruction error signals that the system has deviated from the learned distribution, indicating a potential fault or attack. While effective in stationary environments, standard autoencoders do not inherently account for multi-mode operations. A global autoencoder trained on all modes simultaneously may struggle to learn the fine-grained nuances of specific products, resulting in an overgeneralized model that lacks sensitivity to early-stage faults \cite{xiao2023fault}. This creates a distinct gap in the current literature: the need for a diagnostic framework that is both data-driven (to handle non-linear process complexities) and mode-aware to adapt to changing production schedules. Integrating mode identification with anomaly detection is essential to reduce false positive rates and provide actionable diagnostics in multi-product facilities \cite{quinones2019data}. This research utilizes the Tennessee Eastman Process (TEP) as a benchmark to validate such an approach \cite{reinartz2021extended}. The TEP is a realistic simulation of a complex chemical plant that has been widely used in the process control community for over three decades \cite{yin2012comparison}. Its multi-unit, multi-product nature makes it an ideal proxy for evaluating monitoring systems under diverse operating conditions.

In this work, we propose a hybrid framework to verify the existence and cause of the identified vulnerability. We compare a Global-Agnostic baseline with a Product-Aware counterpart to demonstrate that the observed degradation in attack scenarios stems from the absence of product awareness. The specific objectives of this study are as follows:

\begin{itemize}
    \item To theoretically formulate the ``Blind Spot'' hypothesis, positing that global anomaly detectors implicitly expand their acceptance regions to accommodate multi-mode variance, thereby masking subtle anomalies.
    \item To design and implement a Product-Aware Autoencoder framework that constrains the learning domain to specific product grades, enhancing diagnostic sensitivity.
    \item To quantify the performance advantage of the proposed framework against a global baseline using the Extended Tennessee Eastman Process, focusing on F1-score and ROC-AUC improvements.
    \item To rigorously validate the Blind Spot hypothesis through a ``transition stress test,'' leveraging transition data as a controlled proxy to assess whether the Global-Agnostic model masks attack-like anomalies due to its widened acceptance region, in contrast to the Product-Aware model.

\end{itemize}

The remainder of this paper is organized as follows: Section II surveys the relevant literature on anomaly detection, mode-aware diagnostics, and the limitations of current data-driven approaches. Section III defines the proposed methodology, detailing the theoretical framework for Global versus Product-Aware topologies. Section IV presents the Case Study, describing the Tennessee Eastman Process benchmark, the data partitioning strategy, and the experimental configuration. Section V discusses the empirical results, including the comparative metrics and the analysis of reconstruction error dynamics during mode transitions. Section VI outlines future research directions, and Section VII concludes the paper with a summary of the findings and directions for future research.

\section{Background and Related Work}

To contextualize the proposed framework, it is necessary to understand the convergence of industrial cybersecurity threats, the complexity of multi-product manufacturing, and the evolution of data-driven monitoring techniques.

\subsection{Key Concepts in Industrial Anomaly Detection}

\begin{itemize}
    \item \textbf{Cybersecurity Threats in ICS:} Modern industrial control systems (ICS) face targeted attacks capable of causing physical damage. Notable examples include the 2014 German steel mill attack, where spear-phishing led to massive furnace damage \cite{alladi2020industrial}, and the 2017 Triton malware incident, which breached safety controllers in a petrochemical plant \cite{gibbs2017triton}. These incidents highlight that multi-product facilities are attractive targets where a single intrusion can disrupt multiple process lines.
    
    \item \textbf{Challenges in Multi-Mode Processes:} A significant complication in manufacturing is that plants operate under multiple modes (e.g., different product grades or recipes). Standard "product-agnostic" anomaly detectors often fail here because normal operating conditions change dramatically between modes \cite{wang2012process}. Consequently, models unaware of the current mode may interpret legitimate product changeovers as anomalies, leading to high false alarm rates \cite{chen2021multimode}.
    
    \item \textbf{Autoencoder-Based Monitoring:} Given the scarcity of labeled attack data, research has pivoted to unsupervised learning, specifically Autoencoders (AEs). AEs train on normal operation data to compress and reconstruct process variables. A spike in reconstruction error signals an anomaly, allowing the system to detect unforeseen faults or cyberattacks without requiring explicit failure signatures \cite{wang2015hidden, han2025time}.
\end{itemize}

\subsection{Summary of Related Works}

Table \ref{tab:lit_review} summarizes key contributions in the field of multi-mode anomaly detection and process monitoring. While various statistical and machine learning methods have been applied, there is a clear trend toward integrating mode-awareness with advanced data-driven techniques.

\begin{table}[ht]
\centering
\caption{Summary of Selected Literature on Multi-Mode Process Monitoring}
\label{tab:lit_review}
\renewcommand{\arraystretch}{1.3}
\begin{tabular}{|p{0.25\linewidth}|p{0.3\linewidth}|p{0.35\linewidth}|}
\hline
\textbf{Study} & \textbf{Methodology} & \textbf{Key Contribution/Domain} \\
\hline
Chen et al. \cite{yang2019multimode} & Robust Dictionary Learning & Developed an anomaly detection method for aluminum electrolysis explicitly accounting for operational phases. \\
\hline
Bakdi et al. \cite{bakdi2019data} & Static \& Dynamic PCA & Combined PCA methods to monitor wind turbines across varying wind conditions and operating modes. \\
\hline
Wang et al. \cite{wang2018multimode} & Local Outlier Factor (LOF) & Applied LOF to the Tennessee Eastman benchmark to handle multiple operating regimes. \\
\hline
Xiao et al. (2023) \cite{xiao2023fault} & Deep Autoencoder (DAE) & Demonstrated that DAEs outperform traditional PCA in detecting faults in the Tennessee Eastman process without manual feature engineering. \\
\hline
Wang et al. \cite{wang2018multi} & Multi-subspace Factor Analysis & Introduced Support Vector Data Description (SVDD) to detect faults across different modes. \\
\hline
\end{tabular}
\end{table}

\begin{figure*}[t]
    \centering
    \includegraphics[width=1.1\textwidth]{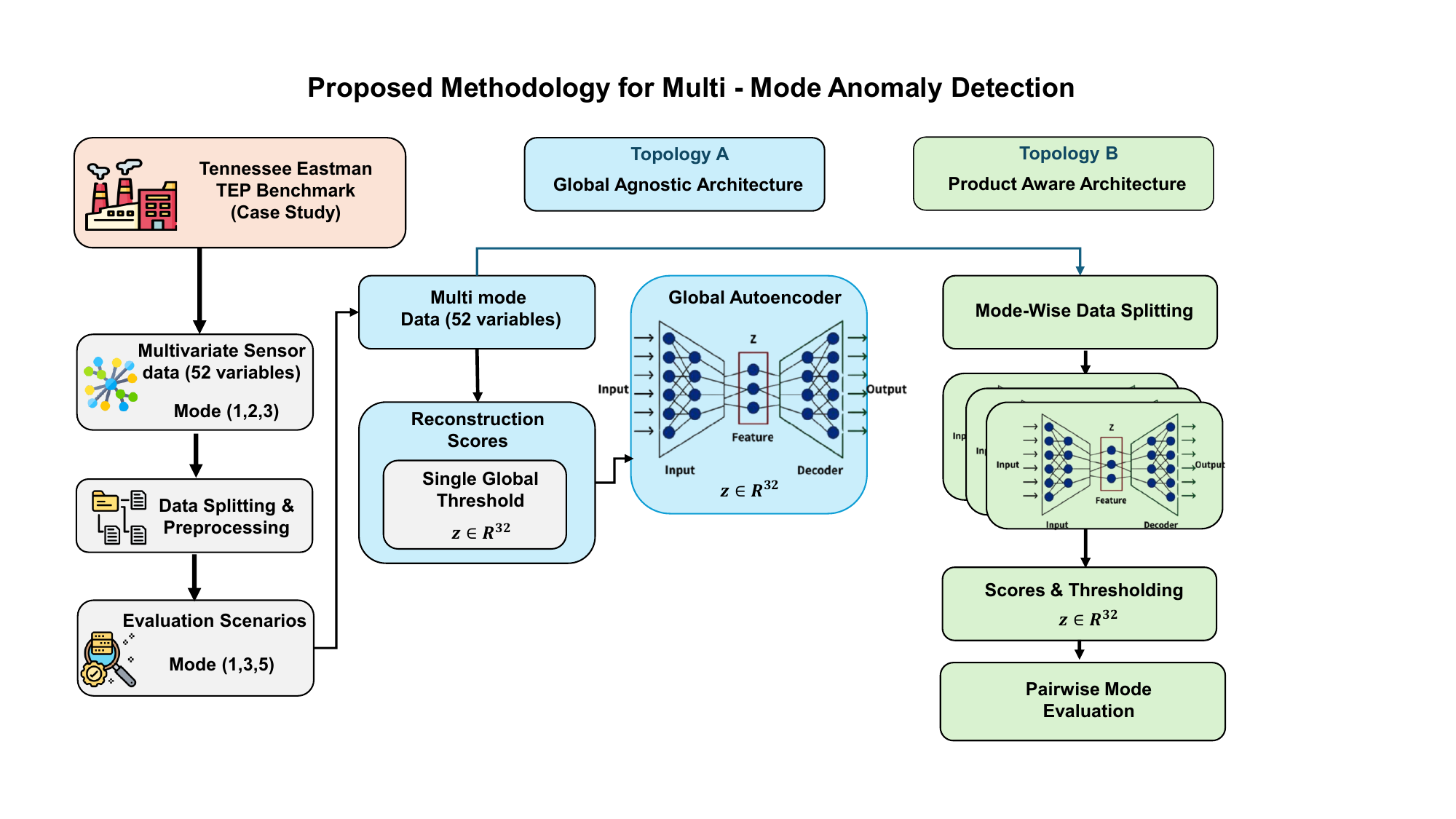} 
    \caption{Comparison of the global vs. product-aware autoencoder architectures. The global model is a single fully-connected autoencoder trained on all modes, whereas the product-aware design trains one autoencoder per mode (1, 3, 5) on mode-specific normal data. All networks share the same symmetric architecture and latent dimension and are optimized using mean-squared reconstruction error with Adam.}
    \label{fig:methodology_pipeline}
\end{figure*}

\subsection{Research Gap and Proposed Methodology}

Despite the advancements listed in Table \ref{tab:lit_review}, a critical gap remains in the intersection of deep learning and mode-aware diagnostics. Current solutions tend to be polarized: they are either \textit{product-agnostic} (ignoring mode differences, resulting in low sensitivity) or \textit{mode-specific} (requiring separate models for every product, which is computationally expensive and difficult to scale) \cite{yang2025uniad}. Furthermore, much of the existing research is confined to specific chemical benchmarks and has not fully addressed the "holistic" need for a system that can handle both the complexity of multiple operating modes and the unpredictability of cyber-physical attacks \cite{chen2021multimode}.

\textbf{Proposed Approach:} To address the trade-off between scalability and diagnostic sensitivity, this work implements a product-aware anomaly detection strategy by training separate autoencoders for each product grade. Each autoencoder learns a grade-specific normal operating manifold from its corresponding data, enabling the detector to preserve sensitivity to subtle faults within that product context. We compare these product-specific models against a global agnostic autoencoder trained on pooled multi-grade data to quantify how global training can mask product-level deviations during multi-mode operations.

\section{Methodology}

In this work, we propose a generalized data-driven framework for anomaly detection in multi-product industrial systems. Unlike traditional statistical process control (SPC) \cite{caulcutt1996statistical} methods that often rely on linear assumptions, our approach leverages deep representation learning to capture non-linear process dynamics. We formally define the challenge of multi-mode operations and synthesize two distinct architectural topologies, Global Agnostic and Product-Aware, to investigate the hypothesis that mode-averaging creates diagnostic ``blind spots'' in the decision boundary.

\subsection{Problem Formulation: Multi-Mode Process Monitoring}

Consider a continuous industrial process equipped with $d$ sensors, producing a state vector $x \in \mathbb{R}^d$ at any time instance $t$. The system operates under a finite set of distinct operational modes (or product grades), denoted as $\mathcal{M} = \{m_1, m_2, \dots, m_K\}$. A fundamental characteristic of such systems is that the statistical distribution of the process variables is conditional on the active mode. That is, for two distinct modes $m_i, m_j \in \mathcal{M}$, the probability density functions differ:
\begin{equation}
    P(x | m_i) \neq P(x | m_j)
\end{equation}
Consequently, a measurement $x$ that is considered nominal (normal) under mode $m_i$ may be statistically anomalous under mode $m_j$, and vice-versa. The objective of the anomaly detection system is to learn a decision function $f(x, m)$ that maps an observation to a binary status $y \in \{0, 1\}$, where $0$ represents normal operation and $1$ represents a fault or cyber-intrusion, conditional on the current operating mode.

\subsection{Deep Autoencoder Framework}

To extract robust features from high-dimensional sensor data without relying on scarce labeled fault data, we employ a Deep Autoencoder (DAE) architecture \cite{farahnakian2018deep}. The DAE functions as a non-linear dimensionality reduction technique, consisting of two parameterized networks: an encoder $E_\phi$ and a decoder $D_\theta$. The encoder maps the high-dimensional input $x$ to a lower-dimensional latent representation $z \in \mathbb{R}^h$ (where $h \ll d$), forcing the network to learn a compressed representation of the significant process correlations:
\begin{equation}
    z = E_\phi(x)
\end{equation}
The decoder subsequently attempts to reconstruct the original input from this latent code:
\begin{equation}
    \hat{x} = D_\theta(z)
\end{equation}
The networks are optimized jointly by minimizing the reconstruction fidelity loss, defined as the Mean Squared Error (MSE) between the input and the output:
\begin{equation}
    \mathcal{L}(\phi, \theta) = \frac{1}{N} \sum_{i=1}^{N} || x_i - D_\theta(E_\phi(x_i)) ||^2
\end{equation}
During inference, the reconstruction error $e(x) = ||x - \hat{x}||^2$ serves as the anomaly score. A low error implies the sample adheres to the learned correlation structure, while a high error indicates a deviation (anomaly).

\subsection{Architectural Topologies: Investigating the Blind Spot}

To address the multi-mode nature of the process, we synthesize and contrast two distinct training paradigms. The divergence in how these topologies construct the latent manifold forms the basis of our ``Blind Spot'' hypothesis.


\subsubsection{\textbf{Topology A: Global Agnostic Architecture}}
The Global architecture represents the standard industrial approach where a single unified model, $\mathcal{M}_{Global}$, is trained on the union of all normal operational data, regardless of the active mode:
\begin{equation}
    \mathcal{D}_{train}^{Global} = \bigcup_{k=1}^{K} \mathcal{D}_{normal}^{(m_k)}
\end{equation}
In this topology, the autoencoder must learn a latent representation that encompasses the variance of all product grades simultaneously. We hypothesize that to minimize the global loss across divergent modes, the model learns a ``loose'' approximation of the process manifold. This effectively expands the volume of the accepted normal space, potentially creating \textit{blind spots}, regions in the latent space where subtle anomalies are masked because they fall within the wide variance allowed for mode transitions.

\subsubsection{\textbf{Topology B: Product-Aware (Mode-Specific) Framework}}
In contrast, the Product-Aware framework treats each operational mode as a distinct domain. We construct a bank of specialized estimators $\{\mathcal{M}_1, \dots, \mathcal{M}_K\}$, where each model $\mathcal{M}_k$ is optimized exclusively on the data distribution of mode $m_k$:
\begin{equation}
    \mathcal{D}_{train}^{(k)} = \{ x \in \mathcal{X} \mid \text{mode}(x) = m_k \}
\end{equation}
By partitioning the problem, each autoencoder $\mathcal{M}_k$ learns a tighter decision boundary specific to the physics of that product grade. This topology requires an auxiliary input or a mode-identification step to route the sensor data to the correct model during inference.

\subsection{Adaptive Thresholding Strategy}

A critical component of the methodology is the conversion of the continuous reconstruction error into a binary decision. A static threshold is insufficient for multi-mode processes due to varying noise levels across product grades. We implement a non-parametric, data-driven thresholding strategy. For the Global architecture, a single scalar threshold $\tau_{global}$ is derived from the $95^{th}$ percentile of the reconstruction errors over the entire global validation set. For the Product-Aware framework, we synthesize a dynamic threshold function $\tau(m)$. For each mode-specific model $\mathcal{M}_k$, a unique threshold $\tau_k$ is computed based on the error distribution of that specific mode:
\begin{equation}
    \tau_k = \text{Percentile}_{95}(\{ e(x) \mid x \in \mathcal{D}_{val}^{(k)} \})
\end{equation}
During online monitoring, the detection logic checks the reconstruction error against the threshold corresponding to the active mode. An anomaly is flagged if and only if $e(x) > \tau_{current}$. This adaptive approach ensures that the detection sensitivity is normalized across different operating regimes, preventing high-variance modes from triggering false alarms while maintaining sensitivity in low-variance modes.

\section{Case Study: The Tennessee Eastman Process}

To validate the proposed framework and quantify the ``blind spot'' phenomenon, we implemented the methodology on the Tennessee Eastman Process (TEP) \cite{reinartz2021extended}. TEP is a realistic simulation of a complex chemical plant that has served as the de facto benchmark for process control and monitoring studies for three decades.

\subsection{Benchmark System Description}

We utilize the Extended TEP simulator provided by Reinartz et al. \cite{reinartz2021extended}, which allows for the simulation of diverse operating regimes. The process consists of five major unit operations: a reactor, a product condenser, a vapor-liquid separator, a recycle compressor, and a product stripper. It involves four gaseous reactants (A, C, D, E) producing two liquid products (G, H) and a byproduct (F). For this study, we focus on three distinct operational modes: Mode 1, Mode 3, and Mode 5, selected to represent significant shifts in production strategy:
\begin{itemize}
    \item \textbf{Mode 1:} Base case operation (50/50 G/H mass ratio).
    \item \textbf{Mode 3:} High production of Product G (90/10 G/H mass ratio).
    \item \textbf{Mode 5:} High production of Product H (10/90 G/H mass ratio).
\end{itemize}
These modes exhibit distinct statistical properties in terms of flow rates, reactor pressures, and temperature profiles, making them an ideal testbed for evaluating product-aware anomaly detection. The dataset includes 52 continuous process variables (sampled every 3 minutes) encompassing reactor levels, component concentrations, and controller outputs.

\subsection{Data Generation and Partitioning}
Data generation was performed using the standard sampling rate of 3 minutes per observation. To ensure a rigorous evaluation, we employed a stratified sampling strategy for training and testing, as detailed in Table \ref{tab:dataset}.

\textbf{Normal Operation Data:} Unlike studies that utilize the pre-fault period of fault runs (which may contain transient instabilities), we extracted ``dedicated normal'' data from stable setpoint variation runs (SpVariation, Magnitude 100). This ensures the training baseline represents a pristine steady-state environment. 

\textbf{Fault Data:} We utilized the standard 28 fault types (IDV1–IDV28) provided by the extended simulator. To evaluate detection robustness, fault samples were stratified across all three modes.

\textbf{Preprocessing:}
We partition the normal data into a training set $\mathcal{D}_{train}$ (66\%) and a testing set $\mathcal{D}_{test}$ (33\%).
\begin{itemize}
    \item The \textbf{Global Model} is trained on the union of training data from Modes 1, 3, and 5.
    \item The \textbf{Product-Aware Models} are trained exclusively on their respective mode-specific partitions.
\end{itemize}
All data is standardized using Z-score normalization. Crucially, the scaler statistics ($\mu, \sigma$) are computed solely on the training split to prevent information leakage into the test set.

\begin{table}[ht]
\centering
\caption{Experimental Dataset Statistics}
\label{tab:dataset}
\begin{tabular}{lcc}
\toprule
\textbf{Data Type} & \textbf{Source} & \textbf{Split Usage} \\
\midrule
Normal Operation & SpVariation (Mag 100) & Train (2/3) / Test (1/3) \\
Fault Types 1--28 & Simulation Completed & Test Only \\
Transition Data & Mode Transitions & Validation Only \\
\midrule
\textbf{Total Input Features} & \multicolumn{2}{c}{52 Continuous Variables} \\
\textbf{Operational Modes} & \multicolumn{2}{c}{Mode 1, Mode 3, Mode 5} \\
\bottomrule
\end{tabular}
\end{table}

\subsection{Model Implementation and Training}

The proposed autoencoder framework was implemented using the PyTorch library \cite{imambi2021pytorch}. The architecture is designed as a symmetric feedforward neural network, specifically engineered to force a dimensionality reduction that captures the non-linear correlations between process variables while filtering out high-frequency sensor noise.
\begin{figure}
    \centering
    \includegraphics[width=1\linewidth]{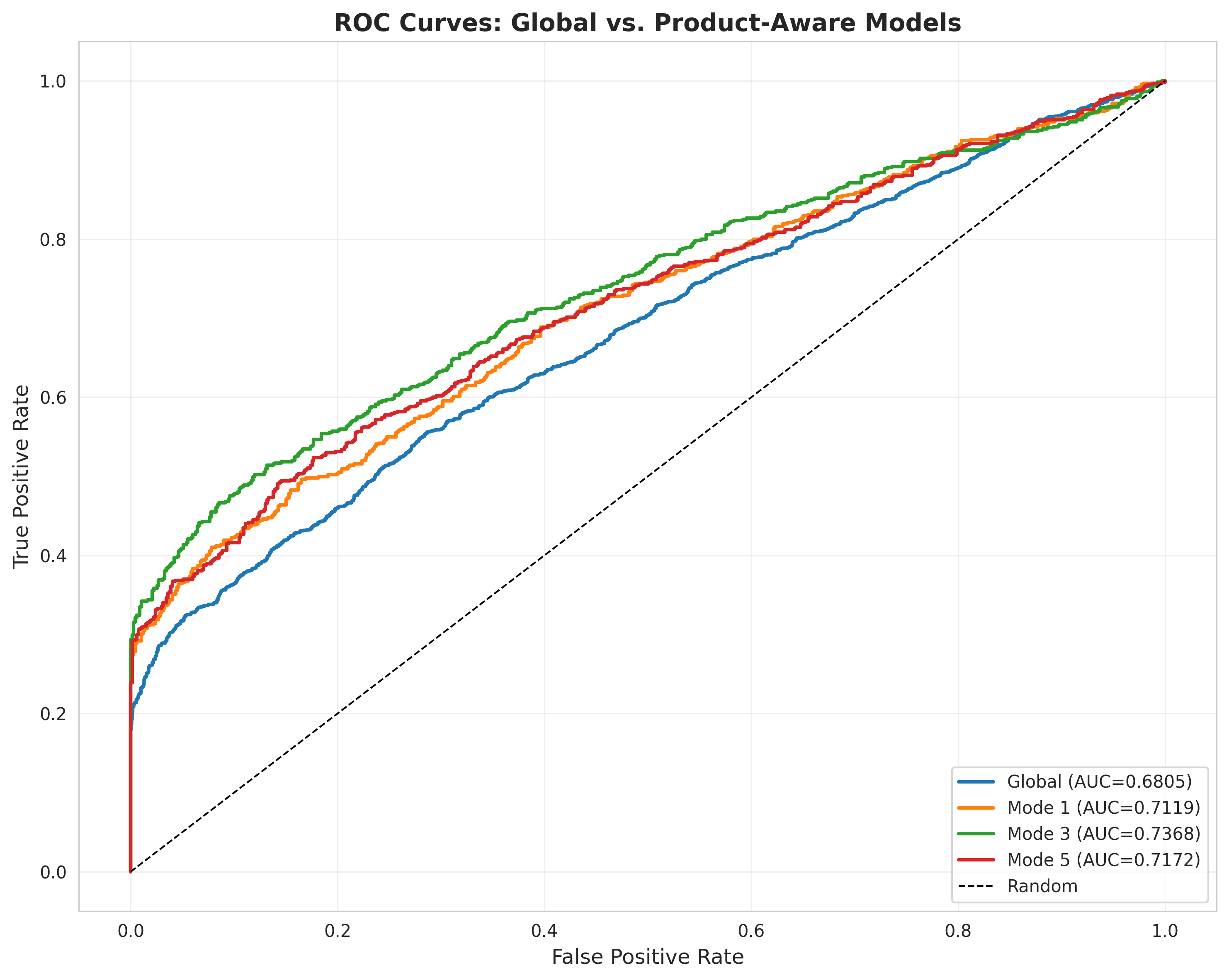}
    \caption{Comparison of Receiver Operating Characteristic (ROC) curves for the Global Agnostic model (blue) versus the Product-Aware mode-specific models (Mode 1, 3, and 5). The Product-Aware curves consistently envelope the Global baseline, indicating that for any given False Positive Rate, the mode-specific models achieve a higher True Positive Rate (sensitivity), thereby validating the superior discriminative power of the constrained learning domains.}
    \label{fig:roc_curves}
\end{figure}
\begin{figure*}[t]  
    \centering
    \includegraphics[width=\textwidth]{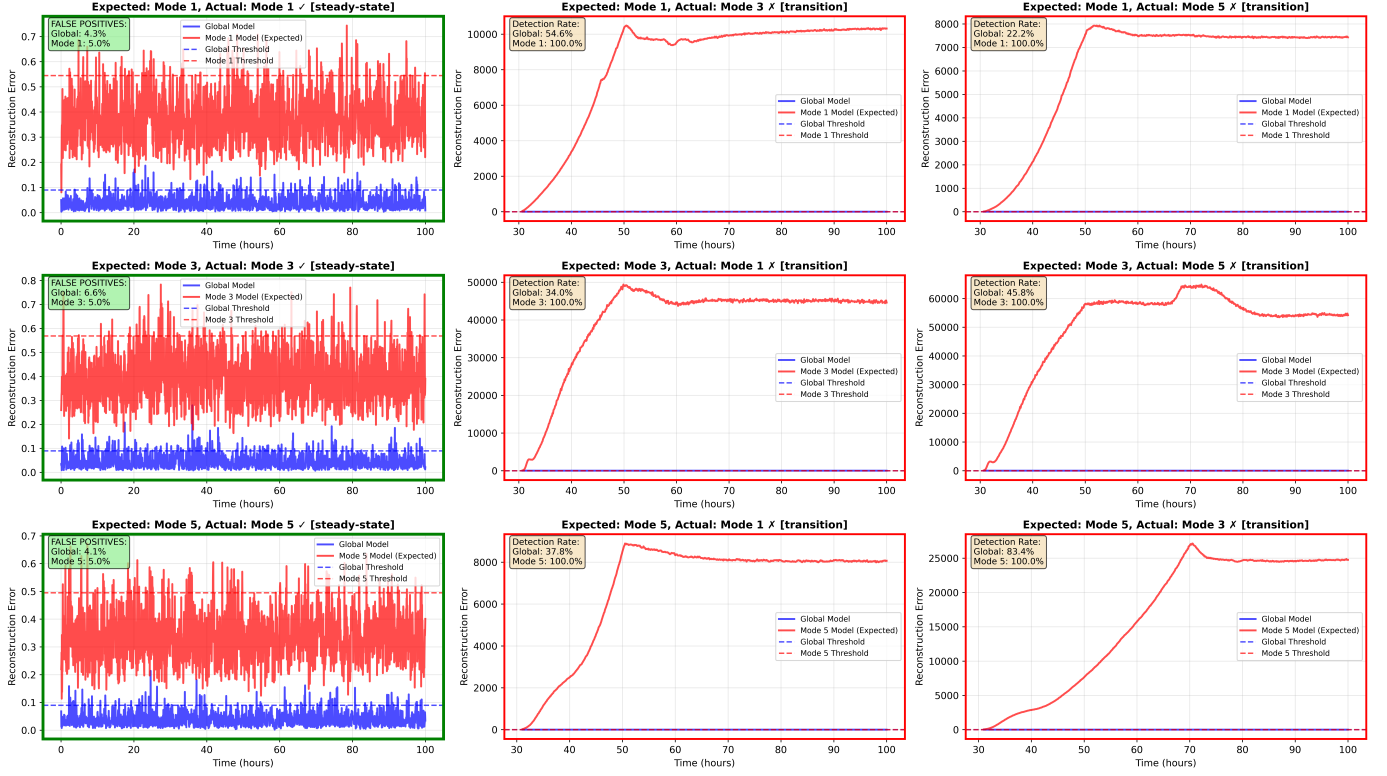}
    \caption{Reconstruction error dynamics during steady-state operation (Left Column) versus unannounced mode transitions (Middle and Right Columns). \textbf{Green boxes} indicate correct operation where both models remain stable. \textbf{Red boxes} indicate the ``Blind Spot'' scenarios: while the Product-Aware model (Red line) registers a massive error spike ($>10,000$) identifying the anomaly, the Global model (Blue line) frequently remains near or below the threshold, failing to detect that the process physics have shifted to a different mode.}
    \label{fig:error_time}
\end{figure*}

\subsubsection{Network Architecture}
We selected a bottleneck topology identified through a hyperparameter grid search, balancing the trade-off between compression ratio and reconstruction fidelity.
\begin{itemize}
    \item \textbf{Encoder ($E_\phi$):} The encoder compresses the 52-dimensional input vector through two dense hidden layers of 128 and 64 units, respectively, culminating in a 32-dimensional latent bottleneck ($z \in \mathbb{R}^{32}$). This bottleneck forces the model to learn a compact representation of the underlying process physics rather than performing an identity mapping.
    \item \textbf{Decoder ($D_\theta$):} The decoder mirrors the encoder structure ($32 \to 64 \to 128 \to 52$), reconstructing the input features from the latent embedding.
    \item \textbf{Layer Components:} Each linear transformation (except the final output layer) is followed by a Batch Normalization layer to stabilize the internal covariate shift and accelerate convergence. We employ Rectified Linear Units (ReLU) as the activation function to mitigate the vanishing gradient problem in deeper layers.
    \item \textbf{Regularization:} To prevent the model from overfitting to specific noise patterns in the training data, which is critical for robust anomaly detection, a Dropout rate of $p=0.2$ is applied after each hidden layer.
\end{itemize}

\subsubsection{Training Dynamics}
The models were trained to minimize the Mean Squared Error (MSE) between the input vector $x$ and the reconstruction $\hat{x}$. The optimization was performed using the Adam algorithm with an initial learning rate of $1e^{-3}$ and a batch size of 256.  To ensure convergence to a robust local minimum, we implemented a dynamic learning rate scheduler. The scheduler monitors the training loss and reduces the learning rate by a factor of 0.5 if the loss fails to improve for 5 consecutive epochs (patience). Training was conducted for a maximum of 50 epochs, which empirical testing showed was sufficient for the loss to stabilize without entering the regime of overfitting.

\subsection{Evaluation Scenarios}

To rigorously test the ``Blind Spot'' hypothesis, we designed two specific evaluation scenarios:

\subsubsection{Scenario A: Steady-State Fault Detection}
In this baseline scenario, we introduce standard process faults (e.g., IDV1: A/C Feed Ratio Step) while the system is in a stable operating mode. We compare the Global model's ability to detect these faults against the Product-Aware models using Area Under the Receiver Operating Characteristic curve (ROC-AUC) and F1-score metrics.

\subsubsection{Scenario B: Unannounced Mode Transitions}
This scenario represents the core of our investigation. We simulate a production changeover (e.g., transitioning from producing Product G in Mode 3 to Product H in Mode 5). 

In a real-world setting, the anomaly detector might not be immediately informed of the schedule change. We evaluate how the Global model (which has learned a wide, averaged distribution) behaves during this transition compared to a Product-Aware system. We hypothesize that the Global model will exhibit a ``dampened'' response, failing to flag the statistical shift as significant, whereas the Product-Aware model, utilizing mode-specific thresholds, will correctly identify the deviation from the previous product grade, thereby offering superior diagnostic resolution.

\section{Results and Discussion}

In this section, we present the empirical evaluation of the proposed frameworks on the Extended Tennessee Eastman Process. We first analyze the general fault detection capabilities across standard process faults (IDV1--28), followed by a targeted investigation into the hypothesized ``blind spots'' during unannounced mode transitions.

\subsection{Comparative Analysis of Detection Metrics}

To establish a baseline, we evaluated both the Global Agnostic model and the Product-Aware ensemble on the hold-out test set, which contains samples from all 28 fault types stratified across Modes 1, 3, and 5. Table \ref{tab:performance_metrics} summarizes the aggregate performance.

\begin{table}[ht]
\centering
\caption{Aggregate Performance Comparison (Test Set)}
\label{tab:performance_metrics}
\begin{tabular}{lccc}
\toprule
\textbf{Metric} & \textbf{Global Model} & \textbf{Product-Aware (Avg)} & \textbf{Improvement} \\
\midrule
F1-Score & 0.4643 & 0.5322 & \textbf{+14.62\%} \\
ROC-AUC & 0.6805 & 0.7220 & \textbf{+6.10\%} \\
\bottomrule
\end{tabular}
\end{table}

The quantitative results indicate a distinct advantage for the Product-Aware topology. While the Global model achieves a respectable ROC-AUC of 0.680, demonstrating that it has learned meaningful process correlations, it is consistently outperformed by the Product-Aware framework, which achieves an average ROC-AUC of 0.722. More significantly, we observe a substantial 14.6\% improvement in the F1-Score (0.464 vs. 0.532). The F1-score is particularly critical in industrial anomaly detection as it balances sensitivity (Recall) with the False Alarm Rate (Precision). The Global model, forced to generalize across divergent operating modes (e.g., the high production rates of Mode 1 vs. the low rates of Mode 3), effectively learns a ``loose'' decision boundary. This results in a lower precision, as the model accepts a wider variance of process behavior as normal. By constraining the training domain to specific products, the Product-Aware models learn tighter, more discriminative boundaries, leading to the observed performance gains.

Figure \ref{fig:roc_curves} (ROC Curves) reinforces this finding. The Product-Aware curves (represented by Modes 1, 3, and 5) consistently envelope the Global model's curve (blue), indicating superior True Positive Rates at equivalent False Positive thresholds.

\begin{figure*}[t]
    \centering
    \includegraphics[width=\textwidth]{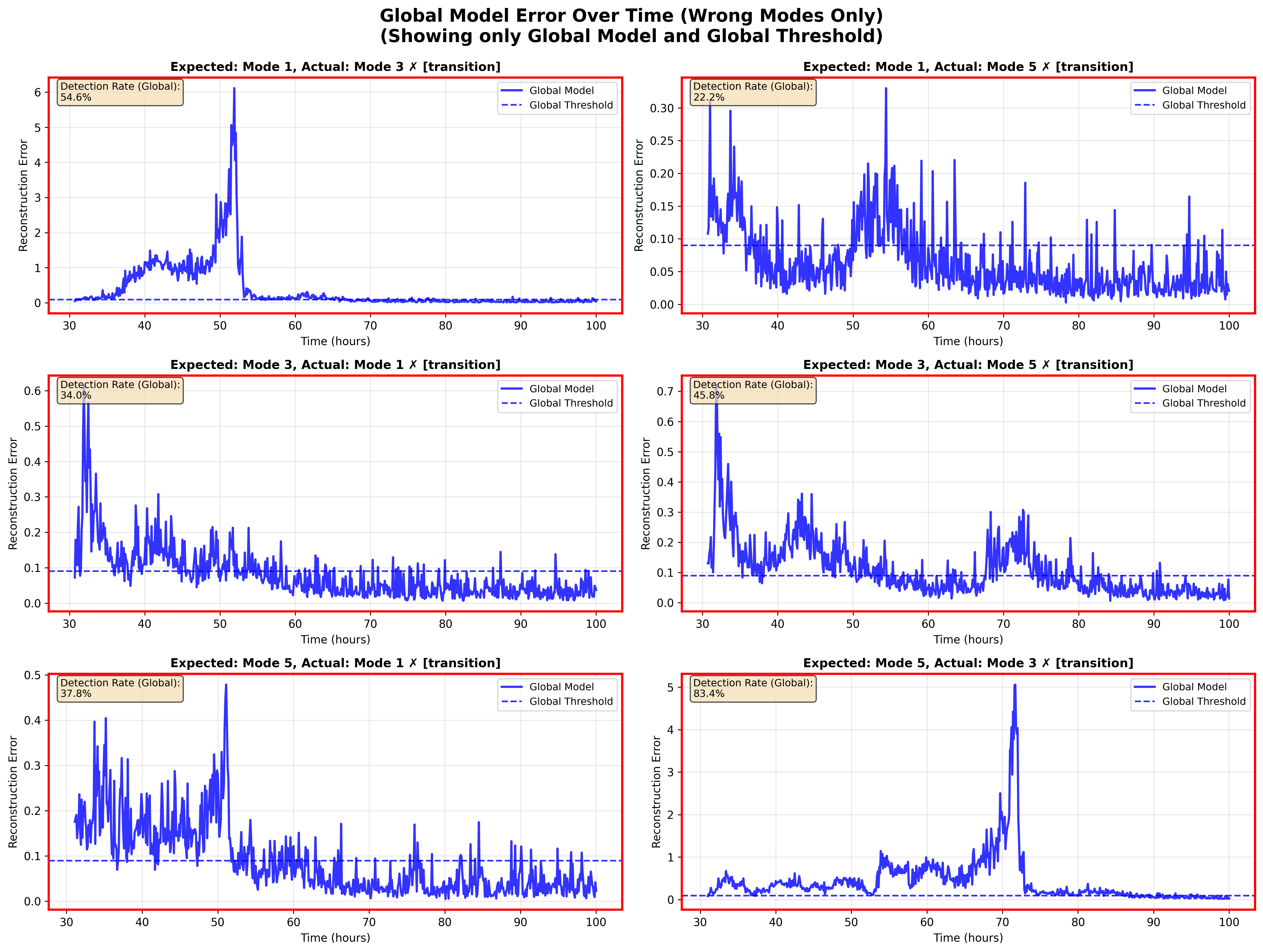}
    \caption{Detailed analysis of the Global Agnostic model's failure modes ("Blind Spots") during unannounced transitions. Each panel displays the reconstruction error (blue trace) when the process shifts to an incorrect operating mode (e.g., Mode 1 $\rightarrow$ Mode 5). The dashed line represents the global alarm threshold ($\tau_{global}$). Despite the process operating in a completely wrong regime, the global model frequently maintains an error score below the threshold (detection rates as low as 22.2\%), illustrating that the model has learned to accept the statistical properties of multiple modes as valid, thereby masking the anomaly.}
    \label{fig:global_error_dynamics}
\end{figure*}

\subsection{Investigation of the Blind Spot: Mode Transitions}

The primary hypothesis of this study is that Global models exhibit ``blind spots'', an inability to detect anomalies that resemble valid operating conditions of \textit{other} products. To test this, we simulated an ``unannounced mode transition'' scenario. In this test, the process transitions (e.g., from Mode 1 to Mode 3) without an accompanying update to the monitoring system. A robust system should flag this deviation immediately.

The results of this stress test provided the strongest evidence for our hypothesis. As illustrated in the error dynamics analysis (Figure \ref{fig:error_time}), we observed a critical divergence in behavior:

\begin{itemize}
    \item \textbf{Product-Aware Response (Correct Detection):} When the process drifted from Mode 1 to Mode 3, the Mode 1-specific autoencoder immediately registered a massive spike in reconstruction error (reaching magnitudes $>10,000$). The detection rate for these events was \textbf{100\%}. The model correctly identified that the physics of the process no longer matched the learned profile of Product 1.
    
    \item \textbf{Global Model Response (The Blind Spot):} Conversely, the Global model failed to consistently flag these transitions. For example, during a transition from Mode 1 to Mode 5, the Global model maintained a detection rate of only \textbf{22.2\%}. Because the Global model was trained on \textit{both} Mode 1 and Mode 5 simultaneously, its latent space creates a continuous valid manifold connecting these operating points. Consequently, the transition data, which represents a severe anomaly for the specific product line, was interpreted by the Global model as acceptable variations within its learned distribution.
\end{itemize}

\subsection{Reconstruction Error Dynamics}

To visualize the mechanics of this failure, we plotted the reconstruction error over time during these transition events. The distinct behaviors are visualized in the error distributions:

\begin{enumerate}
    \item \textbf{Steady-State Stability (Green Boxes):} During valid operation (Left Column), both architectures maintain low reconstruction errors below their respective thresholds ($\tau_{95}$), confirming that the Product-Aware approach does not introduce instability during normal production.
    
    \item \textbf{Transitional Failure (Red Boxes):} The Middle and Right columns illustrate the ``Blind Spot.'' In the top-right plot (Expected Mode 1, Actual Mode 5), the Global model's error (blue line) remains flat and significantly below the alarm threshold. It effectively ``sleepwalks'' through the mode change. In contrast, the Product-Aware model (red line) exhibits an exponential rise in error the moment the process variables deviate from the Mode 1 setpoints.
\end{enumerate}

These results empirically demonstrate that, within the studied benchmark, Product-agnostic anomaly detectors exhibit a structural vulnerability.

\section{Future Work}

While this work establishes the efficacy of product-aware topologies, several avenues remain for extending the framework toward fully autonomous industrial deployment. The current framework relies on the availability of explicit product labels to route data. Future research should focus on "closing the loop" by developing a hierarchical Mode-Supervisor network. This upstream classifier would utilize unsupervised clustering or time-series classification to identify the active operational regime in real-time, dynamically routing sensor streams to the appropriate product-specific autoencoder without human intervention. Detecting an anomaly is only the first step; diagnosis is the second. Future iterations of this work will integrate Explainable AI (XAI) techniques, such as reconstruction error contribution plots or gradient-based attribution methods. By quantifying which specific sensors contribute most to the residual spike, the system could provide operators with actionable insights into the physical location of the fault, significantly reducing the time-to-mitigation. Finally, industrial processes are rarely static; sensors degrade and catalysts age, leading to concept drift. Validating this framework on physical testbeds beyond simulation benchmarks is imperative. Future studies should explore techniques for Continual Learning, allowing the mode-specific models to adapt to slow-moving environmental changes without catastrophic forgetting of the core process physics.

\section{Conclusion}

As industrial systems increasingly integrate complex, multi-product manufacturing lines, the assumption that a single "product-agnostic" anomaly detector can suffice is becoming a critical vulnerability. In this work, we formalized and empirically examined the “Blind Spot” hypothesis on the Extended Tennessee Eastman benchmark, the theory that training a global model on diverse operating modes forces the learner to expand its acceptance region, thereby masking significant process deviations. By implementing a rigorous comparative framework on the Extended Tennessee Eastman Process (TEP), we demonstrated that this vulnerability is not merely theoretical but quantifiable. Our Product-Aware framework, which constrains the learning domain to specific product grades, achieved a 14.6\% improvement in F1-score and a 6.1\% increase in ROC-AUC compared to the standard Global Agnostic baseline. These metrics confirm that specialized estimators capture the non-linear correlations of process physics with significantly higher fidelity than a generalized model.

Most critically, our stress tests on unannounced mode transitions revealed the structural danger of global modeling. When the process shifted between distinct operational modes, the Global model failed to consistently flag the event, exhibiting detection rates as low as 22.2\% (2 out of 9 transition runs detected) in transition scenarios. In contrast, the Product-Aware models achieved 100\% detection, correctly identifying that the process had deviated from the expected thermodynamic profile of the active product. This research indicates that, although global models offer deployment simplicity, they can introduce meaningful security risks in flexible manufacturing environments. In our experiments, the overgeneralized model often sacrificed the sensitivity required to detect mode-related deviations, particularly during unannounced transitions, whereas product-aware estimators preserved this sensitivity.

\bibliographystyle{IEEEtran}
\bibliography{bib}

\end{document}